\documentclass{article}

\PassOptionsToPackage{numbers, compress}{natbib}

\usepackage[preprint]{neurips_2025}

\usepackage[utf8]{inputenc} 
\usepackage[T1]{fontenc}
\usepackage{hyperref}
\usepackage{tabularx}
\usepackage{url}
\usepackage{booktabs}
\usepackage{amsfonts}
\usepackage{nicefrac}
\usepackage{microtype}  
\usepackage{xcolor, colortbl} 
\usepackage{wrapfig}
\usepackage{subcaption}
\usepackage{adjustbox}
\usepackage{graphicx}
\usepackage{amsmath}
\definecolor{lightpink}{rgb}{0.945, 0.816, 0.804}
\definecolor{lightgreen}{rgb}{0.851, 0.906, 0.839}
\definecolor{lightblue}{rgb}{0.8, 0.9, 1}
\definecolor{lightyellow}{rgb}{0.992, 0.949, 0.816}
\usepackage{algorithm}
\usepackage{algorithmic}
\usepackage{pifont}
\newcommand{\cmark}{\ding{51}} 
\newcommand{\xmark}{\ding{55}} 
\usepackage{multirow}
\usepackage{array}
\definecolor{citepcolor}{HTML}{0071BC}
\definecolor{linkcolor}{HTML}{ED1C24}
\usepackage{tcolorbox}
\usepackage{amsmath, amssymb, amsthm}
\usepackage{amsfonts} 

\definecolor{HeaderBlue}{HTML}{E0E8F0} 
\definecolor{RowGray}{HTML}{F5F5F5}

\theoremstyle{remark}

\title{CAPO: Reinforcing Consistent Reasoning in Medical Decision-Making}
\author{%
  Songtao Jiang\textsuperscript{1}\thanks{Equal contribution.} , 
  Yuan Wang\textsuperscript{1}\footnotemark[1],
  Ruizhe Chen\textsuperscript{1}\footnotemark[1],
   Yan Zhang\textsuperscript{1}, Ruilin Luo\textsuperscript{2}, Bohan Lei\textsuperscript{1},\\ 
  \textbf{Sibo Song\textsuperscript{3}, Yang Feng\textsuperscript{4},  Jian Wu\textsuperscript{1}, Jimeng Sun\textsuperscript{5}, Zuozhu Liu\textsuperscript{1}\thanks{Corresponding author.}} \\
  \textsuperscript{1}Zhejiang University \textsuperscript{2}Tsinghua University \textsuperscript{3}Alibaba Group \textsuperscript{4} Angelalign Inc. \textsuperscript{5}UIUC \\ 
}
\begin{document}

\maketitle

\begin{abstract}

In medical visual question answering (Med-VQA), achieving accurate responses relies on three critical steps: precise perception of medical imaging data, logical reasoning grounded in visual input and textual questions, and  coherent answer derivation from the reasoning process. Recent advances in general vision-language models (VLMs) show that large-scale reinforcement learning (RL) could significantly enhance both reasoning capabilities and overall model performance. However, their application in medical domains is hindered by two fundamental challenges: 1) misalignment between perceptual understanding and reasoning stages, and 2) inconsistency between reasoning pathways and answer generation, both compounded by the scarcity of high-quality medical datasets for effective large-scale RL.  In this paper, we first introduce \texttt{Med-Zero-17K}, a curated dataset for pure RL-based training, encompassing over 30 medical image modalities and 24 clinical tasks. Moreover, we propose a novel large-scale RL framework for Med-VLMs, Consistency-Aware Preference Optimization (CAPO), which integrates rewards to ensure fidelity between perception and reasoning, consistency in reasoning-to-answer derivation, and rule-based accuracy for final responses. Extensive experiments on both in-domain and out-of-domain scenarios demonstrate the superiority of our method over strong VLM baselines, showcasing strong generalization capability to 3D Med-VQA benchmarks and R1-like training paradigms. Code and dataset will be released. 


\end{abstract}

\section{Introduction}
Recent advancements in medical decision-making have significantly improved the accuracy of Medical Visual Question Answering (Med-VQA)~\citep{jiang2024med,li2024llava,kim2024mdagents,jiang2025omnivmedscalingmedicalvisionlanguage,jiang2025hscrhierarchicalselfcontrastiverewarding}. However, the high-stakes nature of the medical domain has precipitated an urgent shift in focus towards the interpretability and transparency of the reasoning processes underlying these decisions~\citep{sox2024medical,bouazizi2024enhancing}. In general-purpose domains, Reinforcement Learning (RL) algorithms have proven effective in enhancing model reasoning capabilities~\citep{li2025limrrlscaling,yue2025doesreinforcementlearningreally,jiang2024modality}, concurrently boosting accuracy while eliciting explicit thought processes. Notably, rule-based RL approaches, such as that employed by Deepseek-R1~\citep{deepseekai2025deepseekr1incentivizingreasoningcapability}, exhibit considerable potential due to their ability to foster robust reasoning without relying on extensive, high-quality Supervised Fine-Tuning (SFT) with Chain-of-Thought (CoT) data~\citep{wang2024drivecotintegratingchainofthoughtreasoning,zheng2023ddcotdutydistinctchainofthoughtprompting,10.1145/3581783.3611898,NEURIPS2023_4ec43957}. This characteristic renders them a cost-efficient and particularly promising avenue for the medical domain, where such specialized reasoning data is inherently scarce~\citep{liu2025generalist,liu2024medcotmedicalchainthought,chen2024huatuogpto1medicalcomplexreasoning,zhang2025medrlvremergingmedicalreasoning}.

\begin{figure}
\centering
\includegraphics[width=0.8\linewidth]{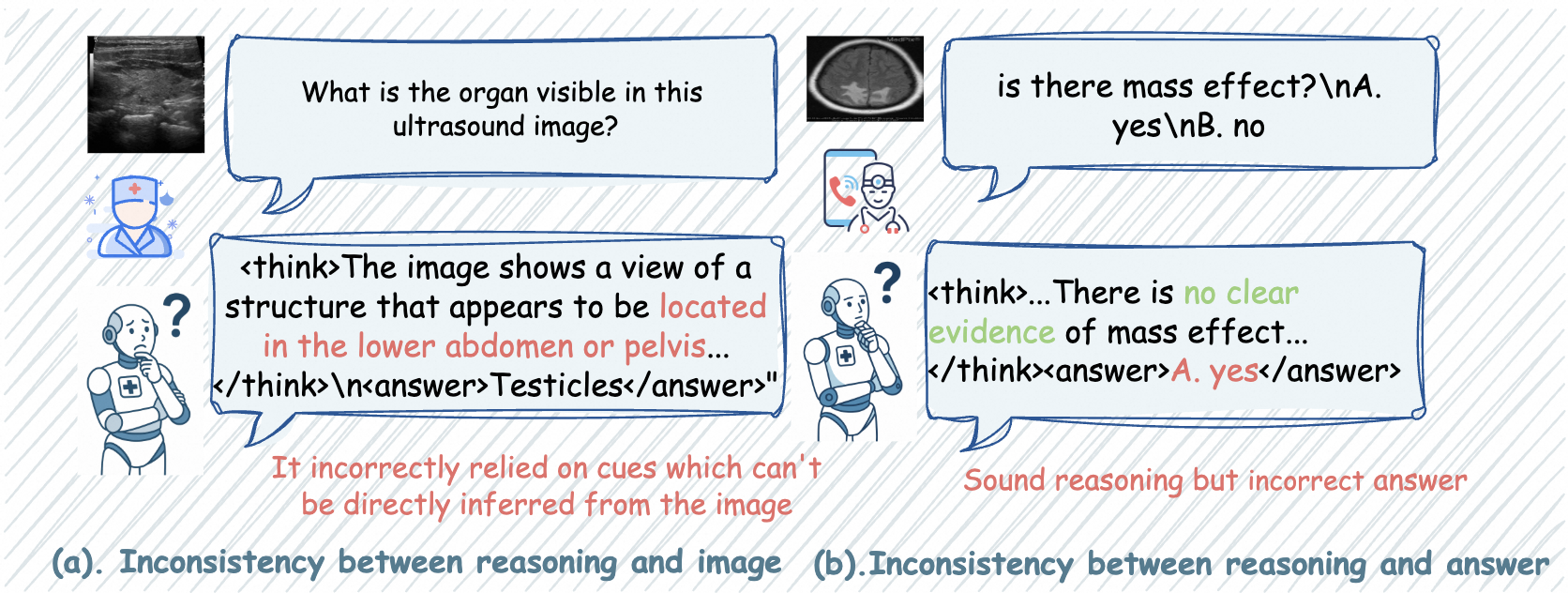}
\caption{Inconsistency patterns in medical decision-making reasoning.}
\label{fig:fig1}
\end{figure}

Despite its promise, recent attempts to directly adapt RL-Zero-style (pure large-scale RL without preliminary SFT) training paradigms from general domains to the medical domain have yielded only limited improvements on established medical benchmarks~\citep{song2025fastcurlcurriculumreinforcementlearning,liu2025understandingr1zeroliketrainingcritical}. Our preliminary investigation identifies two primary reasons. First, we observed a critical issue of reasoning shortcuts: base models pre-trained on general-purpose data and subsequently fine-tuned using RL-Zero-like methods in the medical domain often generate reasoning processes misaligned with the actual visual content of medical images (see Figure~\ref{fig:fig1}). We attribute this to a distributional shift—general-purpose Vision-Language Models (VLMs), with limited exposure to medical images during pretraining, tend to overly rely on the parametric knowledge embedded in their Large Language Model (LLM) backbones. This leads to spurious, text-biased rationales that lack fidelity to medical images, resulting in unreliable predictions. Second, we identified a critical inconsistency between reasoning pathways and answer generation, i.e., the final answer might not adhere to the reasoning processing, leaving the reasoning process ineffective. These issues are exacerbated by the scarcity of high-quality medical datasets for effective large-scale RL training. 

\begin{figure}
\centering
\includegraphics[width=1\linewidth]{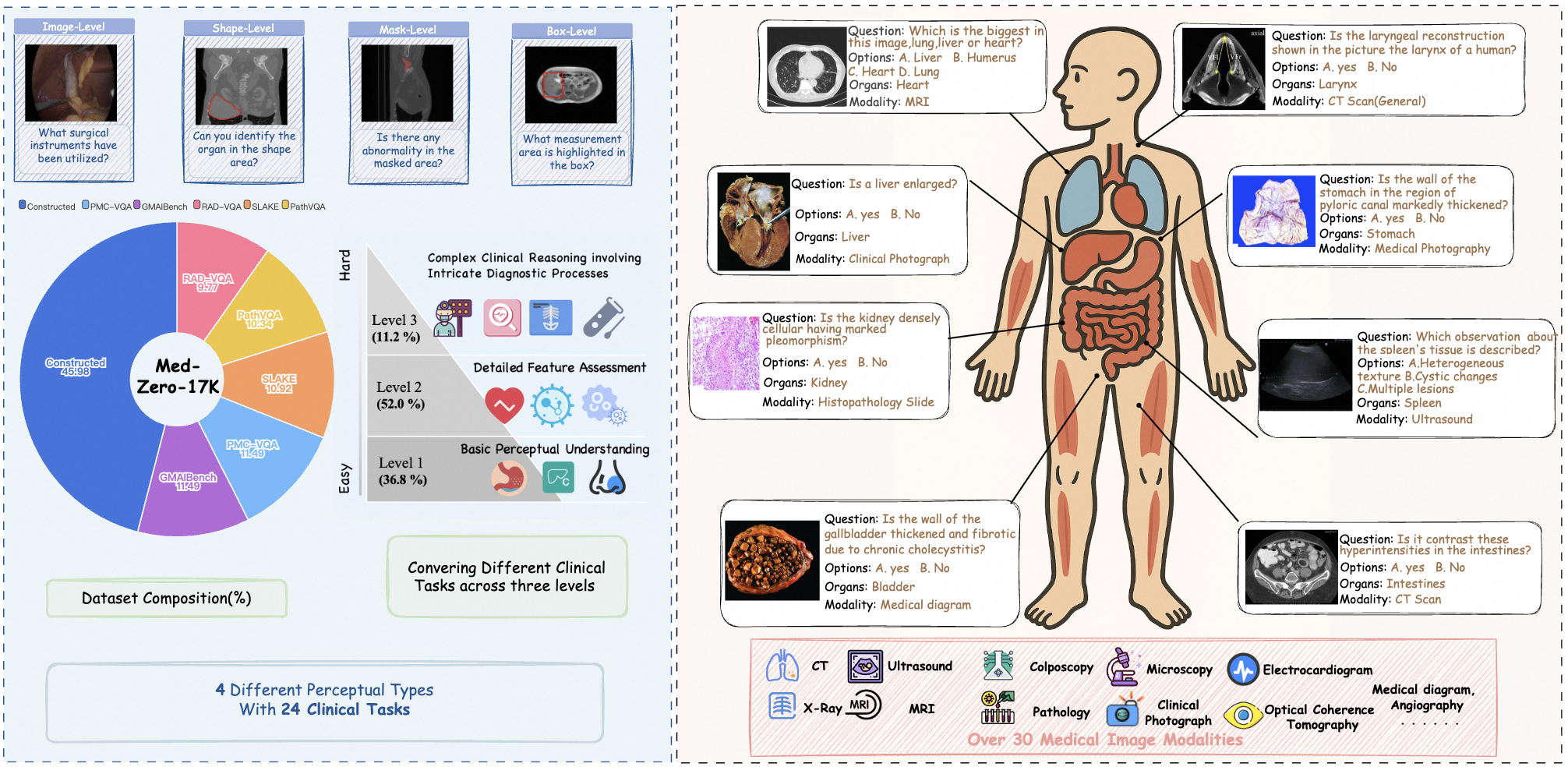}
\caption{Overview of the \texttt{Med-Zero-17K} which encompasses 4 distinct granularity levels of image annotations and 24 diverse clinical tasks, covering various human organs and over 30 modalities.}
\label{fig:pipeline}
\end{figure}

To address these challenges, we develop a novel training framework that jointly advances data and algorithmic designs. At the data level, we introduce \texttt{Med-Zero-17K}, a large-scale and diverse dataset spanning over 30 medical modalities. Unlike prior datasets with narrow task coverage, \texttt{Med-Zero-17K} captures a broad spectrum of real-world clinical scenarios, featuring four levels of annotation granularity and 24 distinct medical tasks, as shown in Figure~\ref{fig:pipeline}.
At the algorithmic level, we propose Consistency-Aware Policy Optimization (CAPO), a new RL framework with verifiable rewards to mitigate the reasoning shortcuts. CAPO integrates two novel consistency rewards besides accuracy-based rewards. 
The perceptual-consistency rewards are obtained by perturbing input medical images to expose text-biased reasoning process, where the model is explicitly rewarded for preferring correct reasoning under original images over perturbed ones, thereby encouraging image-consistent, trustworthy inference. The decision-consistency rewards leverage an external LLM as judge to evaluate the consistency between the reasoning process and the answer.

Extensive experiments across both in-distribution and out-of-distribution datasets demonstrate that our CAPO framework achieves state-of-the-art results, outperforming both domain-specific and general-purpose VLMs. Notably, compared to conventional GRPO, our method achieves more stable and higher rewards, while revealing an interesting finding: \textit{the consistency-related rewards introduced by CAPO encourage longer and deeper clinical reasoning patterns.}
Unlike conventional SFT which suffers from overfitting on datasets such as PMC-VQA~\cite{zhang2023pmc}, CAPO consistently delivers stable and robust improvements under zero-shot conditions. Beyond in-domain gains, CAPO exhibits strong generalization capabilities to unseen modalities and tasks. Without accessing overlapping training data, our approach surpasses both baseline models and SFT methods on the OmnimedVQA~\cite{hu2024omnimedvqa} and MMMU Health \& Medicine benchmarks~\cite{yue2024mmmu}, covering a diverse spectrum of clinical modalities and expert-annotated tasks. These results underscore CAPO's effectiveness in enabling trustworthy and generalizable medical visual reasoning.

    



\section{Related Work}
\paragraph{Reinforcement Learning with Verified Rewards}
RL with Verifiable Rewards (RLVR) has shown strong promise in enabling human-like reasoning without extensive supervision \cite{shao2024deepseekmathpushinglimitsmathematical,lambert2025tulu3pushingfrontiers,luo2025ursaunderstandingverifyingchainofthought}. Traditional methods such as PPO \cite{schulman2017proximalpolicyoptimizationalgorithms} face limitations due to their reliance on reward models, which can suffer from inefficiency and reward hacking \cite{amodei2016concreteproblemsaisafety}. GRPO \cite{shao2024deepseekmathpushinglimitsmathematical} improves on PPO by replacing value functions with group-based scoring, enhancing efficiency and performance in structured reasoning. Extending this, R1-Zero trains models purely via reinforcement learning, guided by rule-based rewards rather than explicit CoT supervision \cite{liu2025understandingr1zeroliketrainingcritical}. This self-evolutionary approach has shown strong generalization and is especially promising for clinical domains where labeled reasoning data is scarce.

\paragraph{Medical Reasoning in Decision-Making}
Earlier efforts in medical decision-making focused on architectural improvements and domain adaptation. LLaVA-Med~\cite{li2024llava} and HuatuoVision~\cite{chen2024huatuogptvisioninjectingmedicalvisual} adapted vision-language models for biomedical understanding; Med-MoE ~\cite{Jiang2024MedMoEMO} used a mixture-of-experts strategy. More recent work emphasizes reasoning capability. Supervised CoT-based methods  \cite{liu2024medcotmedicalchainthought,gai2024medthinkexplainingmedicalvisual,jiang2024joint,jiang2025fastslowintegratingfast} improved transparency but struggled with generalization, while RL-based approaches like Med-R1 \cite{lai2025medr1reinforcementlearninggeneralizable} and MedVLM-R1 \cite{pan2025medvlmr1incentivizingmedicalreasoning} used rule-based rewards to guide reasoning, yet largely relied on in-domain data and failed to generalize across benchmarks. While the R1-Zero paradigm offers task-agnostic scalability, its integration into medical VLMs for developing more robust and generalizable clinical reasoning remains limited~\cite{jayaraman2024primer}.

\begin{figure}
\centering
\includegraphics[width=0.8\linewidth]{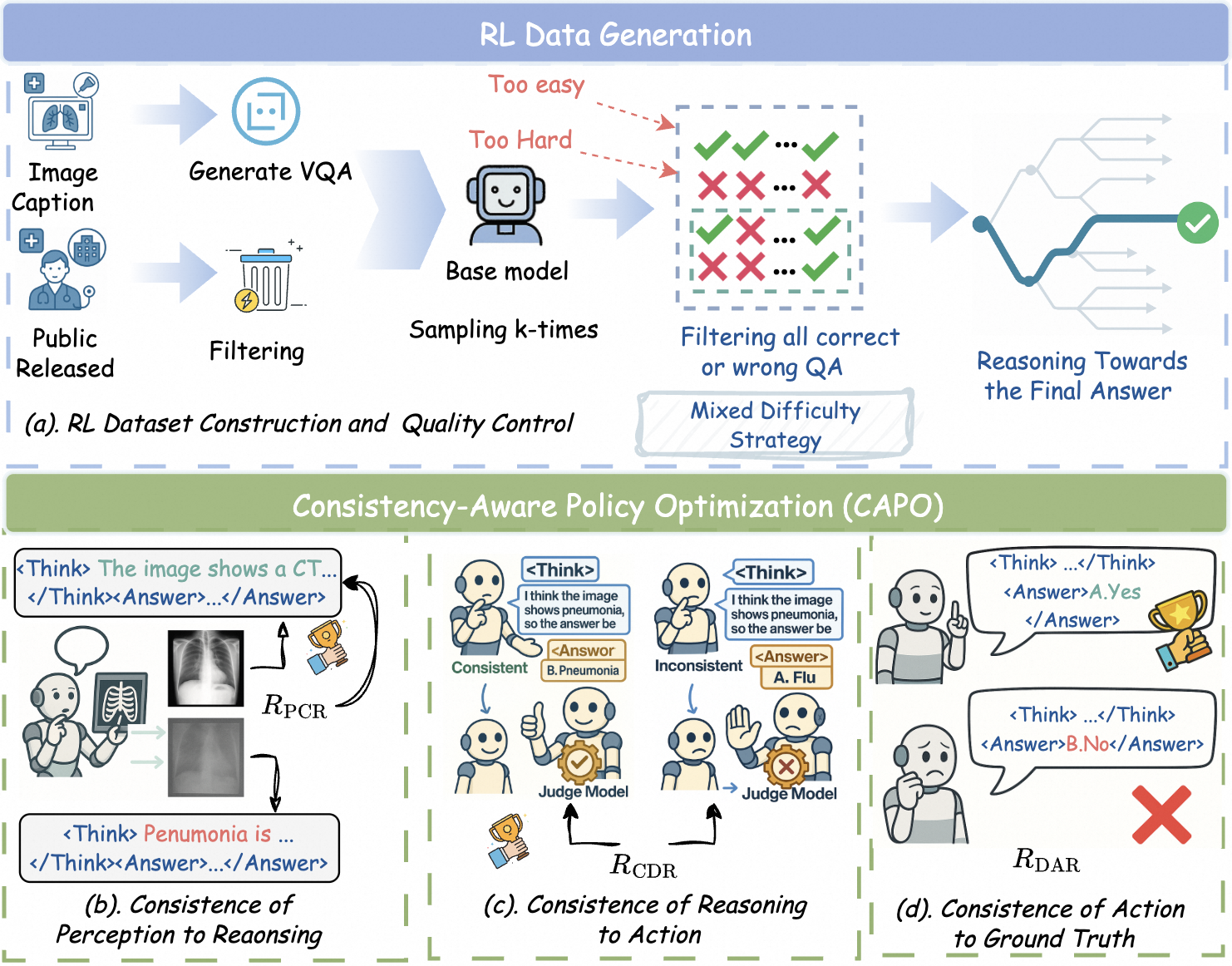}
\caption{Overview of CAPO which comprises four key components: (a) RL dataset construction using a mixed difficulty strategy, (b) $R_{\text{PCR}}$ to align reasoning with visual inputs, (c) $R_{\text{CDR}}$ to maintain coherence between reasoning and decisions, and (d) $R_{\text{DAR}}$ to align decisions with correct answers.}
\label{fig:method}
\end{figure}

\section{Methods}
\label{sec:methods}
\subsection{\texttt{Med-Zero-17K Dataset}}
To address the scarcity of systematically organized, high-quality reinforcement learning (RL) datasets in the medical domain, we introduce \texttt{Med-Zero-17K}, a comprehensive and meticulously curated dataset designed to support RL algorithms. This dataset integrates data from multiple publicly available medical data sources and is enriched through the generation of high-quality VQA data. Specifically, we employed the Qwen-VL-72B~\cite{bai2025qwen2} model to create VQA pairs from Pubmedvision's image-caption data~\cite{chen2024huatuogptvisioninjectingmedicalvisual}, significantly enhancing the dataset's modality diversity and task coverage.
To avoid scenarios where model advantages become skewed~\cite{zhang2017understandingdeeplearningrequires}, we implemented a mixed difficulty data filtering strategy. This involves sampling 10 responses from the base model with a temperature of 0.9, then excluding QA pairs answered entirely correct or entirely incorrect. Retained QA pairs have at least one correct response within these trials, ensuring a dataset exhibiting balanced levels of difficulty.
We further categorized clinical questions into three complexity levels to represent a progression from basic perception (Level 1: foundational visual analysis), through detailed feature assessment (Level 2), to intricate clinical reasoning tasks involving complex diagnostics (Level 3). The specific distribution of these difficulty levels is illustrated in Figure~\ref{fig:pipeline}.

Overall, \texttt{Med-Zero-17K} encompasses 30 distinct medical imaging modalities, including CT, MRI, pathology images, as well as less prevalent imaging sources. It covers diverse anatomical regions and is specifically structured to address 24 clinically relevant tasks. Additionally, annotations are provided across 4 perceptual granularities: image-level, shape-level, bounding box, and segmentation masks. The carefully structured \texttt{Med-Zero-17K} dataset facilitates stable, progressive learning during RL training, empowering models to transition seamlessly from fundamental visual perception to sophisticated clinical reasoning. 

\subsection{Consistency-Aware Policy Optimization (CAPO)}
Traditional policy optimization methods like GRPO~\cite{shao2024deepseekmath} can exhibit inconsistencies between different components of the reasoning process. To address this issue, we introduce Consistency-Aware Policy Optimization (CAPO) as shown in Figure~\ref{fig:method}, a framework that promotes coherence throughout the generation process through specialized reward signals. CAPO incorporates three reward components: a Perceptual-Cognitive Consistency Reward, a Cognitive-Decision Consistency Reward, and a standard Decision Accuracy Reward.

\paragraph{Cognitive-Decision Consistency Reward ($R_{\text{CDR}}$):}
To ensure that the final answer logically follows from the preceding reasoning steps, we introduce $R_{\text{CDR}}$. This component leverages a Large Language Model (LLM) as an external judge. Let $p$ denote the model-generated reasoning process and $a$ be the final answer, where $o = (p,a)$. We define a judgment function $\mathcal{J}(p, a)$ which consults an LLM (specifically, GPT-4o, using the prompt template detailed in Appendix~\ref{tab:prompt_template_consistency}) to evaluate the logical entailment of $a$ by $p$. The reward is then:
\begin{equation}
R_{\text{CDR}}(o_i = <p_i, a_i>) = 
\begin{cases}
r_{\text{cdr}} & \text{if } \mathcal{J}(p_i, a_i) = \text{True} \\
0 & \text{otherwise}
\end{cases}
\label{eq:cdr}
\end{equation}
where $\mathcal{J}(p_i, a_i) = \text{1}$ indicates that the LLM judge deems the reasoning process to logically support the final answer. The hyperparameter $r_{\text{cdr}}$ (e.g., $0.1$) is the reward value for consistency.

\paragraph{Decision Accuracy Reward ($R_{\text{DAR}}$):}
This reward incentivizes correct answers after reasoning:
\begin{equation}
R_{\text{DAR}}(o_i) = 
\begin{cases} 
r_{\text{dar}} & \text{if $o_i$ is correct} \\
0 & \text{if $o_i$ is incorrect}
\end{cases}
\label{eq:dar}
\end{equation}
where $r_{\text{dar}}$ is a hyperparameter reward value (e.g., $0.8$).

\paragraph{Perceptual-Cognitive Consistency Reward ($R_{\text{PCR}}$):}
This reward function incentivizes the reward model to produce reasoning that is more robust to visual perturbations, thereby reducing hallucinations caused by overreliance on language priors, which can lead to incorrect or unreliable reasoning. During reinforcement learning, we generate rollouts using both the original visual inputs $I$ and their corrupted versions $\tilde{I}$. Let $o$ and $\tilde{o}$ denote the model responses to original and corrupted inputs, respectively. The Perception-Consistency Reward ($R_{\text{PCR}}$) is defined as:
\begin{equation}
R_{\text{PCR}}(o_i, \tilde{o}_i) = 
\begin{cases} 
r_{\text{pcr}} & \text{if } \text{R}_{\text{DAR}}(o_i) - \text{R}_{\text{DAR}}(\tilde{o}_i) > \tau_{\text{pcr}} \\
0 & \text{otherwise}
\end{cases}
\label{eq:pcr}
\end{equation}
where $\tau_{\text{pcr}}$ is a predefined threshold (set to $0.3$). $r_{\text{pcr}}$ is the reward value (set to $0.1$ in our experiments).

The total reward for an output $o_i$ under CAPO, denoted $R_{\text{CAPO}}(o_i)$, is the sum of these individual components:
\begin{equation}
R_{\text{CAPO}}(o_i) = R_{\text{PCR}}(o_i) + R_{\text{CDR}}(o_i) + R_{\text{DAR}}(o_i)
\label{eq:capo_reward}
\end{equation}
This aggregated reward $R_{\text{CAPO}}(o_i)$ is used for advantage calculation. For each output $o_i$ in a group of $G$ outputs generated in response to a query $q$, the advantage $A_i$ is computed by normalizing the CAPO rewards within the group, analogous to GRPO:
\begin{equation}
A_i = \frac{R_{\text{CAPO}}(o_i) - \text{mean}(\{R_{\text{CAPO}}(o_j)\}_{j=1}^G)}{\text{std}(\{R_{\text{CAPO}}(o_j)\}_{j=1}^G) + \delta}
\label{eq:advantage}
\end{equation}
The policy is updated by optimizing a clipped surrogate objective with KL regularization:
\begin{equation}
J_{\text{CAPO}}(\theta) = \mathbb{E}_{q, \{o_i\} \sim \pi_{\theta_{\text{old}}}} \left[ \frac{1}{G} \sum_{i=1}^{G} L_{\text{CLIP}}(o_i, q, \theta) \right] - \beta D_{\text{KL}}(\pi_\theta(\cdot|q) \| \pi_{\text{ref}}(\cdot|q))
\label{eq:capo_objective}
\end{equation}
where $L_{\text{CLIP}}(o_i, q, \theta)$: The PPO-style~\cite{schulman2017proximal} clipped surrogate loss for output $o_i$ towards user query $q$, weighted by the normalized CAPO advantage $A_i$. $\beta$ is a hyperparameter controlling the strength of the KL penalty $D_{\text{KL}}$. $\pi_{\text{ref}}$ is a reference policy. By integrating these consistency-promoting rewards, CAPO yields policies that demonstrate not only accuracy but also robust and coherent reasoning processes across the perception, reasoning, and decision-making stages.

\section{Experiments}
\begin{table}[ht!]
\caption{\label{tb-res1} The results of the medical VQA benchmark.}
\centering \small
\begin{tabular}{lcccc|c}
\toprule 
  {Model}& {VQA-RAD} & {SLAKE} & {PathVQA} & {PMC-VQA} & {Avg.} \\ \midrule
{Gemini-2.0-flash-lite} & {59.4} & {73.1} & {64.9} & {50.8} & {60.5} \\
{GPT-4.1-Nano} & {61.8}& {73.1}  & {70.6}& {53.1}  & {64.5} \\ 
Med-Flamingo  & 45.4& 43.5  & 54.7& 23.3  & 41.7 \\
RadFM & 50.6& 34.6  & 38.7& 25.9& 37.5 \\
LLaVA-Med-7B & 51.4& 48.6  & 56.8& 24.7  & 45.4 \\
Qwen-VL-Chat  & 47.0& 56.0  & 55.1& 36.6  & 48.9 \\
Yi-VL-34B  &  53.0& 58.9  & 47.3& 39.5 & 49.7\\
LLaVA-v1.6-7B  & 52.6& 57.9  & 47.9& 35.5  & 48.5 \\
LLaVA-v1.6-13B & 55.8& 58.9  & 51.9& 36.6  & 50.8 \\
LLaVA-v1.6-34B & 58.6& 67.3  & 59.1& 44.4  & 57.4 \\ 
LLaVA-v1.5-LLaMA3-8B  & 54.2& 59.4  & 54.1& 36.4  & 51.0  \\
LLaVA\_Med -LLaMA3-8B& 60.2& 61.2  & 54.5& 46.6& 55.6  \\
{PubMedVision-8B}  & {63.8}& {74.5}  & {59.9}& {52.7}  & {62.7} \\ 
{HuatuoGPT-Vision-34B}  & {68.1}& {76.9}  & {63.5}& {58.2}  & {66.7} \\ 
Qwen2.5-VL-7B & 70.9& 72.8& 65.7  & 54.9  & 66.0\\ 
Qwen2.5-VL-7B + SFT& 75.2& 81.0& 66.9  & 52.2  & 68.8\\ 
\rowcolor{lightblue}Qwen2.5-VL-7B + CAPO& 78.5& 79.1  & 68.9  & 55.5 & 70.5\\ 
\bottomrule
\end{tabular} 
\end{table}

\begin{table}[ht!]
\centering\small
\caption{\label{tb-res2} The accuracy of OmniMedVQA within different modalities. Specifically, {FP} denotes \textit{Fundus Photography}, {MRI} denotes \textit{Magnetic Resonance Imaging}, {CT} denotes \textit{Computed Tomography}, {OCT} denotes \textit{Optical Coherence Tomography}, {Der} denotes \textit{Dermoscopy}, {Mic} denotes \textit{Microscopy Images}, {US} denotes \textit{Ultrasound}, and {X-Ray} denotes \textit{X-Ray}.}
\begin{tabular}{lcccccccc|c}
\toprule
{Model} & {CT} & {FP} & {MRI} & {OCT} & {Der} & {Mic} & {X-Ray} & {US} & {Avg.} \\ \midrule
Med-Flamingo & 34.6 & 33.3 & 27.5 & 26.0 & 28.3 & 28.1 & 30.1 & 33.2 & 30.2 \\
RadFM & 33.3 & 35.0 & 22.0 & 31.3 & 36.3 & 28.0 & 31.5 & 26.1 & 30.5 \\
LLaVA-Med-7B & 25.3 & 48.4 & 35.9 & 42.1 & 45.2 & 44.0 & 31.7 & 83.7 & 44.5 \\
Qwen-VL-Chat & 51.5 & 45.4 & 43.9 & 54.0 & 55.4 & 49.5 & 63.1 & 33.5 & 49.5 \\
Yi-VL-34B & 39.8 & 57.2 & 51.4 & 70.5 & 54.5 & 61.4 & 64.2 & 40.5 & 54.9 \\
LLaVA-v1.6-7B & 40.1 & 39.5 & 54.8 & 58.4 & 54.0 & 48.8 & 53.3 & 47.9 & 49.6 \\
LLaVA-v1.6-13B & 40.0 & 43.6 & 47.4 & 63.2 & 58.0 & 50.5 & 59.6 & 42.6 & 50.6 \\
LLaVA-v1.6-34B & 50.6 & 63.4 & 60.9 & 68.4 & 65.7 & 62.8 & 74.7 & 44.5 & 61.4 \\
LLaVA-v1.5-LLaMA3-8B & 33.0 & 49.7 & 53.8 & 76.0 & 63.1 & 48.4 & 56.6 & 31.2 & 48.8 \\
LLaVA\_Med -LLaMA3-8B& 60.8 & 68.5 & 66.3 & 79.0 & 66.6 & 60.3 & 73.3 & 49.3 & 65.5 \\
{PubMedVision-8B} & 61.6 & 80.2 & 65.1 & 86.3 & 71.6 & 67.4 & 81.4 & 87.4 & 75.1 \\
HuatuoGPT-Vision-34B & 60.8 & 85.5 & 66.5 & 90.0 & 74.0 & 71.3 & 83.8 & 81.7 & 76.7 \\ 
Qwen2.5-VL-7B & 63.9 & 73.3 & 68.5 & 74.0 & 67.1 & 73.9 & 74.7 & 33.4 & 66.1 \\
Qwen2.5-VL-7B + SFT & 62.1 & 66.3 & 62.5 & 59.6 & 59.2 & 66.6 & 74.6 & 34.4 & 60.7 \\ 
\rowcolor{lightblue}Qwen2.5-VL-7B + CAPO & 65.7 & 82.1 & 74.3 & 75.6 & 67.8 & 74.7 & 75.4 & 37.7 & 69.2 \\
\bottomrule
\end{tabular}
\end{table}

\begin{table}[ht!]
\small \centering
\caption{\label{tb-test_mmmu}Results on the MMMU Health \& Medicine track. The Health \& Medicine track is divided into five categories: \textbf{BMS} for \textit{Basic Medical Science}, \textbf{CM} for \textit{Clinical Medicine}, \textbf{DLM} for \textit{Diagnostics and Laboratory Medicine}, \textbf{P} for \textit{Pharmacy}, and \textbf{PH} for \textit{Public Health}. Results are obtained by submitting to the official website.}
\begin{tabular}{lccccc|c}\toprule
{Model} & {BMS} & {CM} & {DLM} & {P} & {PH} & \begin{tabular}[c]{@{}c@{}}MMMU\\ Health \& Medicine\end{tabular} \\ \midrule
{Gemini-2.0-flash-lite} &  56.7 &  66.8 &  43.3 &  66.8 &  60.0 &  58.7 \\
{GPT-4.1-Nano} &  63.3 &  60.0 &  43.3 &  63.3 &  73.3 &  60.6 \\
Med-Flamingo &  29.6 &  28.1 &  24.8 &  25.3 &  31.2 &  28.3  \\ 
RadFM &  27.5 &  26.8 &  25.8 &  24.7 &  29.1 &  27.0  \\
LLaVA-Med-7B &  39.9 &  39.1 &  34.6 &  37.4 &  34.0 &  36.9  \\
Qwen-VL-Chat &  36.5 &  31.7 &  32.7 &  28.4 &  34.6 &  32.7  \\
Yi-VL-34B&  49.4 &  48.9 &  43.2 &  40.5 &  32.0 &  41.5  \\
LLaVA-v1.6-7B  &  40.5 &  36.9 &  32.1 &  32.3 &  26.9 &  33.1  \\
LLaVA-v1.6-13B &  53.6 &  46.7 &  33.3 &  22.2 &  40.0 &  39.3  \\
LLaVA-v1.6-34B &  56.4 &  56.0 &  46.9 &  46.7 &  41.7 &  48.8  \\
LLaVA-v1.5-LLaMA3-8B &  42.3 &  44.0 &  37.0 &  34.7 &  35.2 &  38.2 \\
LLaVA\_Med -LLaMA3-8B&  48.2 &  43.8 &  42.0 &  39.7 &  35.8 &  41.1 \\
{PubMedVision-8B} &  61.0 &  58.8 &  50.0 &  44.7 &  38.7 &  49.1 \\
HuatuoGPT-Vision-34B &  64.6 &  62.5 &  50.6 &  54.1 &  44.2 &  54.4 \\ 
Qwen2.5-VL-7B&  53.6 &  60.0 &  40.0 &  66.7 &  53.3 &  54.7 \\
Qwen2.5-VL-7B + SFT & 57.1 & 60.0 & 26.7 & 55.6 & 60.0 & 51.9 \\ 
\rowcolor{lightblue}Qwen2.5-VL-7B + CAPO&  60.7 &  70.0 &  40.0 &  74.1 &  56.7 &  60.0 \\ 
\bottomrule
\end{tabular}
\end{table}

\subsection{Experiments Setup}
\paragraph{Evaluation Benchmarks.} 
We extensively validate the effectiveness of our method across a wide range of downstream tasks. Following previous work~\cite{chen2024huatuogptvisioninjectingmedicalvisual,li2024llava,jiang2024med}, we conduct comprehensive experiments on six benchmarks across three categories: (i). \textit{Medical Visual Question Answering Benchmark:} We evaluate on the widely-used test sets of Rad-VQA~\cite{lau2018dataset}, SLAKE~\cite{liu2021slake},  PathVQA~\cite{he2020pathvqa}, and PMC-VQA~\cite{zhang2023pmc} to assess capabilities in medical visual question answering. Specifically, for SLAKE, we utilize its English CLOSED segment.
(ii). \textit{Multimodal Benchmark:} We employ the Health \& Medicine track of MMMU~\cite{yue2024mmmu}, a widely-adopted multimodal benchmark that primarily evaluates models' general capabilities in the medical domain. This benchmark is particularly challenging as it contains questions from real clinical settings that demand expertise comparable to medical professionals. Since its questions do not overlap with any training datasets, it serves as an out-of-distribution test for our method.
(iii). \textit{Traditional Medical Imaging Tasks:} We leverage the open-access portion of the OmniMedVQA dataset~\cite{hu2024omnimedvqa}, which encompasses 42 traditional medical imaging datasets. This benchmark allows us to explore our method's performance across diverse medical imaging modalities.

\paragraph{Baselines.} We conducted comprehensive performance comparisons against a wide range of baseline models, which can be categorized into two main classes: (i) \textbf{Medical domain-specific VLMs}, where we selected current state-of-the-art medical vision-language models including HuatuoGPT-Vision-34B~\cite{chen2024huatuogptvisioninjectingmedicalvisual}, Med-Flamingo~\cite{moor2023med}, and RadFM~\cite{wu2023towards}; (ii) \textbf{General-purpose VLMs}, where we compared against leading models such as LLaVA-v1.6-34B~\cite{li2024llavanextinterleavetacklingmultiimagevideo}, Yi-VL-34B~\cite{young2024yi}, and Qwen2.5-VL-72B~\cite{bai2025qwen2}. Through fair comparisons with these two types of SOTA models, we demonstrate the effectiveness of our proposed approach.

\subsection{Main Results}
\paragraph{Consistent In-domain Performance Improvement.}
Our method demonstrates superior in-domain performance, yielding consistent and significant improvements across established medical VQA benchmarks. Leveraging the Qwen2.5-VL-7B backbone, our approach achieves state-of-the-art results on VQA-RAD and PathVQA datasets, outperforming both specialized medical models and general-purpose VLMs. Notably, CAPO exhibits enhanced stability and robustness compared to conventional supervised fine-tuning (SFT), which suffers from performance degradation on the PMC-VQA dataset due to overfitting tendencies. 
\paragraph{Strong Generalization on Out-of-Distribution Tests.}
To rigorously evaluate generalization capabilities, we employed OmnimedVQA and MMMU Health and Medicine benchmarks—datasets whose sources remain disjoint from our training corpus. As evidenced in Table~\ref{tb-res2}, our approach consistently surpasses both the baseline model and SFT across eight diverse medical imaging modalities, including common formats (CT, MRI, X-Ray) and specialized ones (OCT, ultrasound). Furthermore, as demonstrated in Table~\ref{tb-test_mmmu}, CAPO achieves state-of-the-art performance on the expert-annotated MMMU Health and Medicine subset, with consistent gains across various clinical tasks. These results substantiate our method's robust generalization across heterogeneous medical domains and clinical scenarios, highlighting its potential for real-world medical applications.

\begin{table}[htbp] 
\small
\caption{Ablation Study of Key Components on Various VQA Datasets}
\label{tab:ablation_study}
\centering
\begin{tabular}{@{}ccc ccccc@{}}
 \toprule
 $R_{PCR}$ & $R_{CDR}$ & $R_{DAR}$ & VQA-RAD & SLAKE & PathVQA & OmnimedVQA & \begin{tabular}[c]{@{}c@{}}MMMU Health \\ \& Medicine\end{tabular} \\
 \midrule


 \xmark & \xmark & \xmark & 70.9 & 72.8 & 65.7 & 66.1 & 54.7 \\

 \xmark & \xmark & \cmark & 74.1 & 74.8 & 67.2 & 65.4 & 57.2 \\

 \xmark & \cmark & \cmark & \underline{76.9} & \underline{76.7} & \underline{68.1} & 68.1 & \underline{59.7} \\

 \cmark & \xmark & \cmark & 76.5 & 75.5 & 67.5 & \underline{69.4} & 58.4 \\

 \cmark & \cmark & \cmark & \textbf{78.5} & \textbf{79.1} & \textbf{68.9} & \textbf{70.5} & \textbf{60.0} \\

 \bottomrule
\end{tabular}
 \end{table}

\begin{figure}[ht]
\centering
\includegraphics[width=0.8\linewidth]{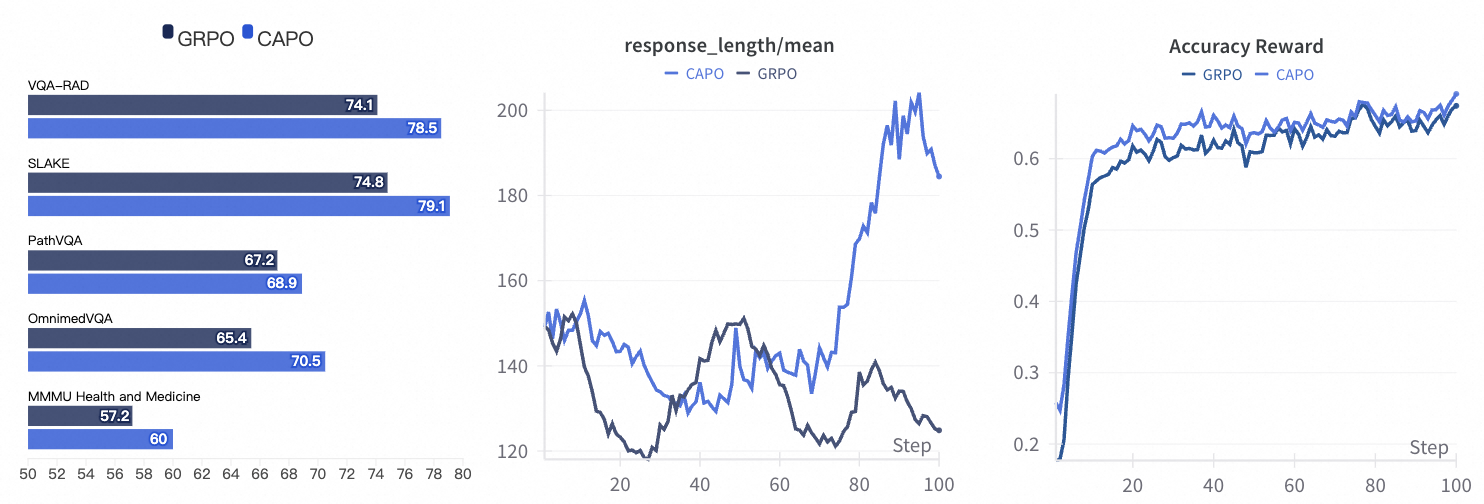}
\caption{Comparision of CAPO and GRPO.}
\label{fig:curve}
\end{figure}

\subsection{Ablation Study and Analysis}
\begin{table*}[!htbp] 
\centering 
\begin{minipage}[t]{0.49\textwidth} 
    \centering 
    \captionof{table}{Ablation on Diffusion Steps} 
    \label{tab:mp_adj_diffusion_steps_full} 
    \setlength{\tabcolsep}{3pt} 
    \begin{adjustbox}{width=\linewidth, center} 
        \begin{tabular}{l|ccc}
        \hline
        \textbf{Diffusion Steps} & \textbf{RADVQA} & \textbf{SLAKE} & \textbf{PathVQA} \\ 
        \hline
        100 (Ours)       & 78.5 & 77.1 & 68.9 \\ 
        0                & 70.9 & 72.8 & 65.7 \\
        300              & 77.2 & 75.5 & 78.3 \\
        \hline
        \end{tabular}
    \end{adjustbox}
\end{minipage}\hfill 
\begin{minipage}[t]{0.49\textwidth} 
    \centering
    \captionof{table}{Ablation on Noise Type}
    \label{tab:mp_adj_noise_type_full}
    \setlength{\tabcolsep}{3pt}
    \begin{adjustbox}{width=\linewidth, center}
        \begin{tabular}{l|ccc}
        \hline
        \textbf{Noise Type}        & \textbf{RADVQA} & \textbf{SLAKE} & \textbf{PathVQA} \\ 
        \hline
        Diffusion (Ours) & 78.5 & 77.1 & 68.9 \\ 
        Mask                   & 76.1 & 74.5 & 66.9 \\
        Crop                   & 77.7 & 76.2 & 68.4 \\
        \hline
        \end{tabular}
    \end{adjustbox}
\end{minipage}
\vspace{\floatsep}
\begin{minipage}[t]{0.49\textwidth} 
    \centering
    \captionof{table}{Ablation on Reward Ratio}
    \label{tab:mp_adj_reward_ratio_full}
    \setlength{\tabcolsep}{3pt}
    \begin{adjustbox}{width=\linewidth, center}
        \begin{tabular}{l|ccc}
        \hline
        \textbf{$R_{\text{DAR}}$:$R_{\text{PCR}}$:$R_{\text{CDR}}$} & \textbf{RADVQA} & \textbf{SLAKE} & \textbf{PathVQA} \\ 
        \hline
        8:1:1 (Ours)      & 78.5 & 77.1 & 68.9 \\ 
        2:1:1             & 72.5 & 73.3 & 67.2 \\
        1:1:1             & 71.7 & 70.9 & 65.4 \\
        \hline
        \end{tabular}
    \end{adjustbox}
\end{minipage}\hfill
\begin{minipage}[t]{0.49\textwidth} 
    \centering
    \captionof{table}{Ablation on Threshold $\tau_{\text{pcr}}$}
    \label{tab:mp_adj_threshold_full}
    \setlength{\tabcolsep}{3pt}
    \begin{adjustbox}{width=\linewidth, center}
        \begin{tabular}{l|ccc}
        \hline
        \textbf{Threshold $\tau_{\text{pcr}}$} & \textbf{RADVQA} & \textbf{SLAKE} & \textbf{PathVQA} \\ 
        \hline
        0.3 (Ours)         & 78.5 & 77.1 & 68.9 \\ 
        0.1                & 76.5 & 75.2 & 66.6 \\
        0.5                & 75.7 & 76.0 & 68.2 \\
        \hline
        \end{tabular}
    \end{adjustbox}
\end{minipage}
\end{table*}

\paragraph{Ablation of Designed Rewards.}
Table~\ref{tab:ablation_study} presents a comprehensive ablation study examining the contribution of each reward component in our framework. The results demonstrate that each individual reward element makes a statistically significant contribution to the overall performance. Moreover, the integration of all designed rewards yields substantial and consistent improvements across diverse medical benchmarks. These findings validate the efficacy of our reward design strategy in enhancing both the robustness and accuracy of medical decision-making.
\begin{figure}[t!]
\centering
\includegraphics[width=0.8\linewidth]{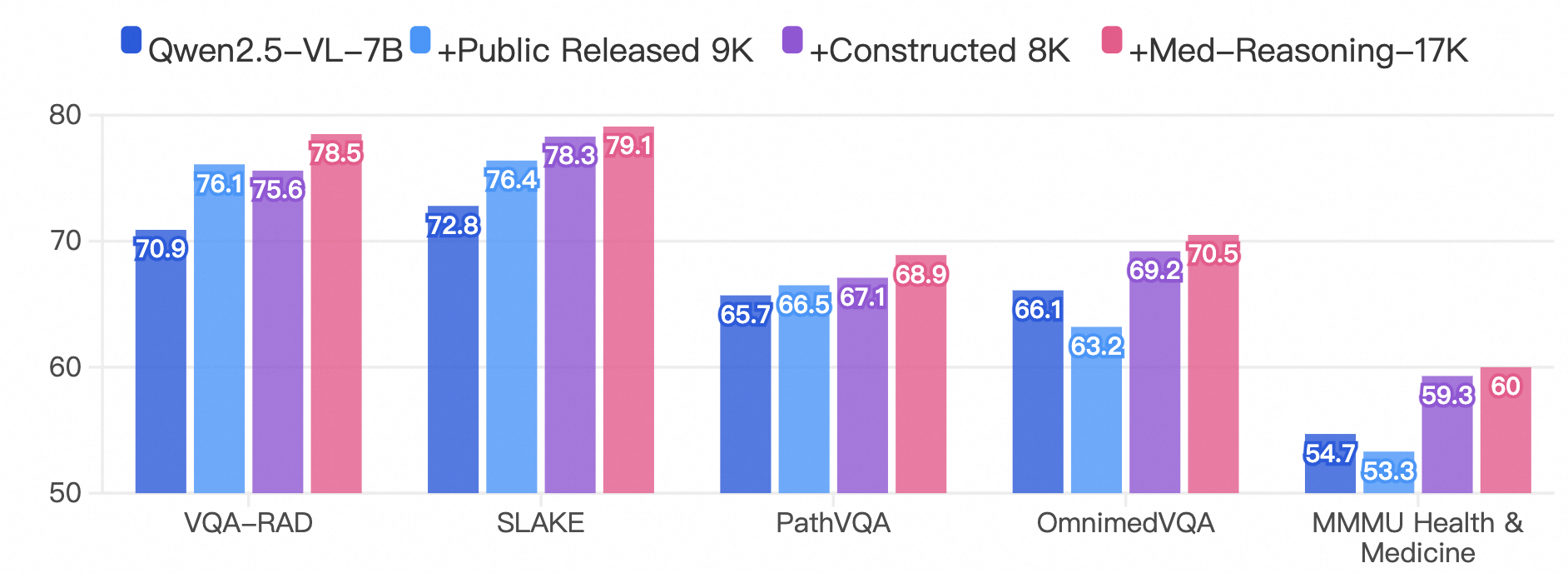}
\caption{Effect of different composition of Med-Zero-17K.}
\label{fig:data_ablation}
\end{figure}

\paragraph{Ablation of Hyperparameters.} 
We conduct comprehensive experiments to validate our hyperparameter choices. 
First, as shown in Table~\ref{tab:mp_adj_diffusion_steps_full}, we investigate the impact of diffusion steps used to generate corrupted images. We observe that excessive steps lead to over-destruction of image content, with 100 steps providing the optimal balance. 
Second, in Table~\ref{tab:mp_adj_noise_type_full} we evaluate various noise types for image corruption and find that diffusion noise outperforms alternatives by introducing sufficient but not excessive corruption, effectively triggering the model's inherent hallucination behaviors. 
Third, our experiments on reward ratios in Table~\ref{tab:mp_adj_reward_ratio_full} reveal that $R_{\text{DAR}}$ contributes most significantly to performance gains, thus deserving the highest proportion, while other rewards play complementary roles. 
Finally, in Table~\ref{tab:mp_adj_threshold_full} we explore the impact of the $\tau_{\text{pcr}}$ parameter in $R_{\text{PCR}}$ and determine that 0.3 represents the optimal value, as both higher and lower values impede reward convergence.

\paragraph{Comparison of CAPO and GRPO.}
As depicted in Figure~\ref{fig:curve}, CAPO consistently achieves higher critic scores and superior VQA performance compared to the GRPO baseline throughout RL training process. We also noticed an intriguing finding from the figure:
\textit{Consistency-based Rewards Promote Deeper and More Stable Reasoning Patterns.}
Specifically, Figure~\ref{fig:curve} illustrates significant fluctuations in response length for GRPO during training, while CAPO exhibits a steady increase in response length. This consistency highlights CAPO's capability to foster stable and sustained reasoning processes. The implication is that consistency-oriented rewards uniquely empower CAPO to enhance cognitive depth, aligning its reasoning closely with input contexts and resulting answers, thus promoting more comprehensive and insightful medical reasoning.

\begin{figure}[t!]
\centering
\includegraphics[width=0.8\linewidth]{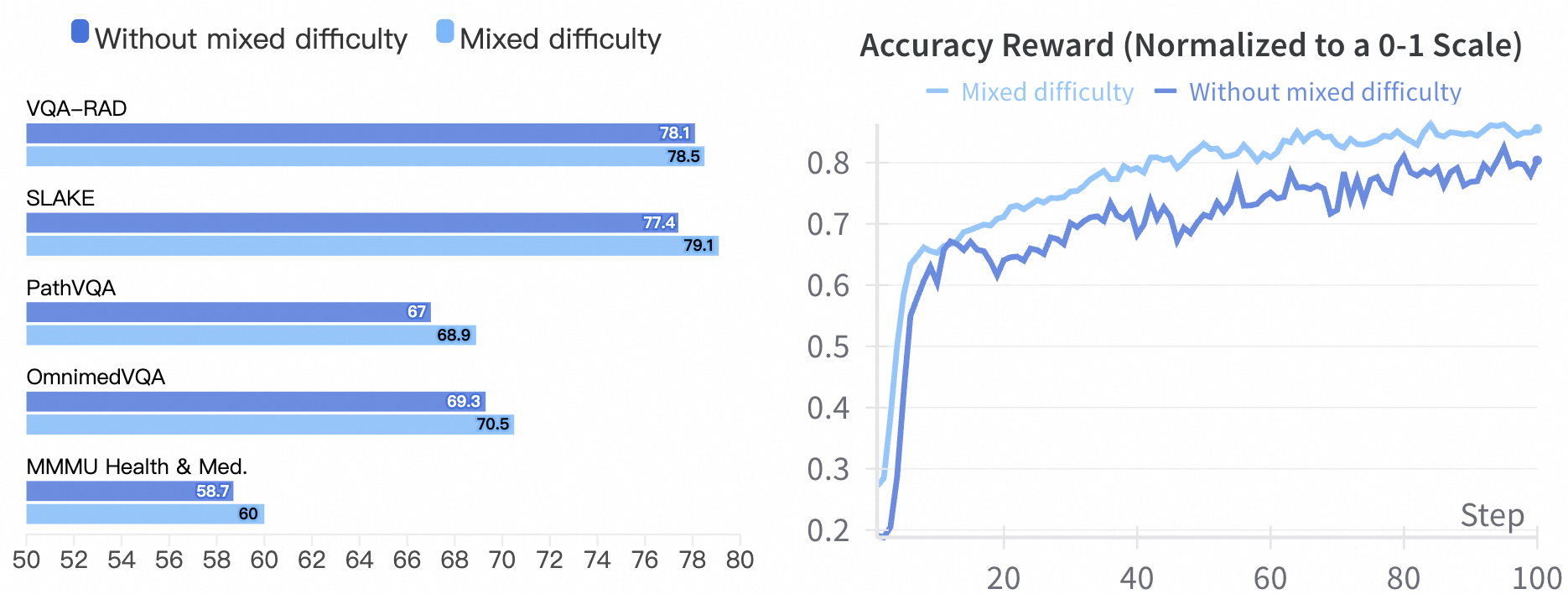}
\caption{Effect of Mixed Difficulty Strategy}
\label{fig:mixed_stra}
\end{figure}
\paragraph{Effect of Med-Zero-17K Composition.}
We conducted a detailed analysis of the composition of Med-Zero-17K. We separately evaluated the impact of the curated 8K high-quality VQA pairs and the cleaned 9K pairs from public released datasets. As shown in Figure~\ref{fig:data_ablation}, both subsets improved performance on RAD-VQA, SLAKE, and PathVQA. However, the publicly sourced 9K pairs struggled to generalize to more diverse and complex benchmarks such as OmnimedVQA and MMMU-Health. This limitation reflects the narrow modality coverage and single-task focus of existing public datasets. In contrast, our VQA pairs from Pubmedvision image-caption data address these shortcomings by significantly boosting the model's generalization across a wide range of modalities and clinical tasks.

\paragraph{Effect of Mixed Difficulty Strategy.}
To investigate the learning dynamics of RL training, we evaluated the effect of incorporating samples with varying difficulty levels. As shown in Figure~\ref{fig:mixed_stra}, our mixed difficulty strategy leads to consistent performance gains across all benchmarks. This improvement stems from alleviating "advantage bias", where models trained on uniformly easy or hard samples tend to produce skewed advantage estimates, leading to unstable or suboptimal policy updates. By balancing samples across the difficulty spectrum, the model receives more nuanced feedback, fostering more reliable and effective learning dynamics.

\paragraph{Generalization from 2D to 3D Medical Images.}

To test the adaptability of our method to 3D contexts, we used M3D~\cite{bai2024m3d} for VQA tasks involving 3D medical images, such as CT scans. As shown in Figure~\ref{fig:3d_results}, our approach outperforms both standard SFT and GRPO baselines when trained on the same RL-Reasoning-17K dataset. These results demonstrate strong generalization across both imaging dimensions and modalities, highlighting the method's potential for broader usage.

\begin{figure}[t!]
\centering
\includegraphics[width=0.9\linewidth]{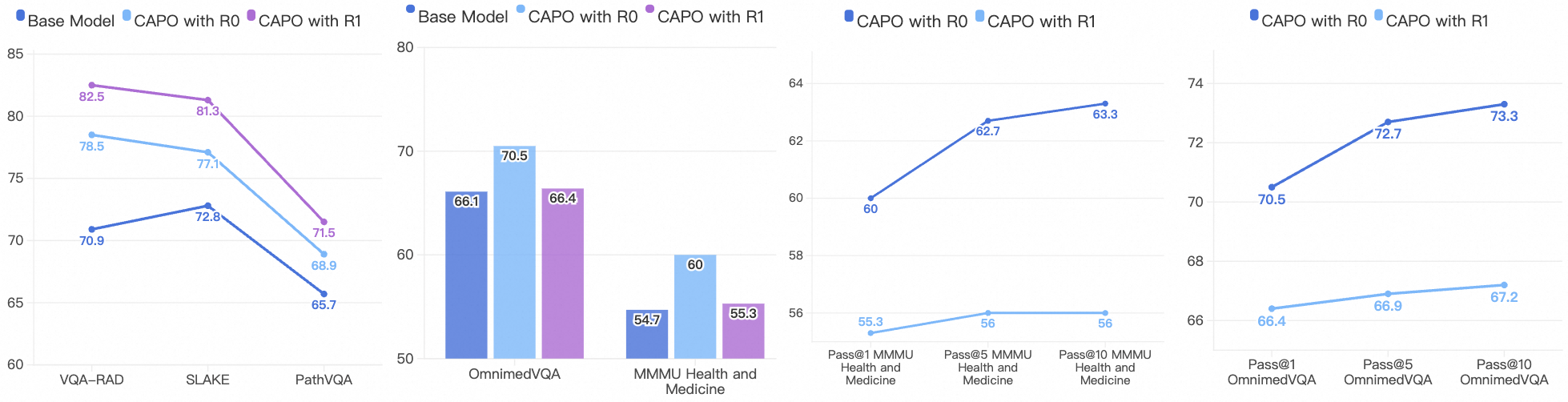}
\caption{Comparision of Different RL Training Paradigms}
\label{fig:r0r1}
\end{figure}
\vspace{-1em}
\paragraph{Are R1-Like Paradigms Effective in the Medical Domain?}
Although R1-like paradigms have proven successful in mathematical reasoning—particularly by cold-starting with CoT data during SFT before RL—their effectiveness in the medical domain remains unclear. To explore this, we used Qwen2.5-VL-72B for rejection sampling to collect high-quality medical CoT data for cold-start SFT(Details in Appendix~\ref{fig:cot}). As shown in Figure~\ref{fig:r0r1}, while the R1 paradigm outperformed the Zero variant on in-domain benchmarks (RAD-VQA, SLAKE, PVQA), it performed significantly worse on OOD tasks. We attribute this to domain mismatch introduced during SFT, which may constrain the model's exploration capacity. Supporting this, our Pass@k evaluation in Figure~\ref{fig:r0r1} shows that the RL-Zero paradigm achieves greater gains across VQA datasets, indicating it better preserves the base model's exploratory abilities. In contrast, R1's cross-domain SFT likely induces overfitting and poor generalization during RL, ultimately limiting its effectiveness in OOD scenarios.

\section{Conclusion}
In this paper, we propose a new RL framework CAPO for Med-VQA that addresses key challenges in aligning perception, reasoning, and answer generation. To support this, we introduce Med-Zero-17K, a diverse dataset spanning over 30 medical modalities and 24 clinical tasks. Experiments on in-domain, out-of-domain, and 3D benchmarks show that CAPO outperforms strong baselines, enabling more consistent and generalizable medical reasoning. Our work demonstrates the potential of consistency-aware RL in advancing medical AI.
\appendix


\bibliography{custom}
\bibliographystyle{unsrt}
\medskip
{
\small


\appendix
\section{Appendix}
 \subsection{Preliminaries} 
\paragraph{Group Relative Policy Optimization (GRPO)}
Group Relative Policy Optimization (GRPO)~\cite{shao2024deepseekmath} is a reinforcement learning algorithm derived from Proximal Policy Optimization (PPO) \cite{schulman2017proximal}, tailored for enhancing model performance, particularly in reasoning tasks, while optimizing resource utilization. A key distinction of GRPO is its departure from the traditional actor-critic PPO structure by forgoing an explicit value function (critic model). Instead, GRPO estimates the necessary baseline for advantage calculation directly from the rewards of a group of outputs. For a given input prompt $q$, the policy model $\pi_{\theta}$ generates a group of $G$ candidate outputs $\{o_1, o_2, \dots, o_G\}$. Each output $o_i$ is then evaluated by a reward model $r_{\varphi}$, yielding rewards $\{r_1, r_2, \dots, r_G\}$.

The advantage $\hat{A}_{i,t}$ for token $o_{i,t}$ in output $o_i$ is computed based on the relative rewards within this group. Specifically, rewards are often normalized by subtracting the group average and dividing by the group standard deviation, i.e., $\tilde{r}_i = (r_i - \text{mean}(\{r_j\}_{j=1}^G)) / \text{std}(\{r_j\}_{j=1}^G)$. For outcome supervision, this normalized reward $\tilde{r}_i$ serves as the advantage for all tokens in output $o_i$. For process supervision, rewards can be assigned per reasoning step, and advantages are typically summed from future step-wise normalized rewards.
The GRPO objective function to be maximized is:
{\small
$$ \mathcal{J}_{\text{GRPO}}(\theta) = \mathbb{E}_{q, \{o_i\} \sim \pi_{\theta_{\text{old}}}} \left[ \sum_{i=1}^{G} \sum_{t=1}^{|o_i|} \left( \min\left(\rho_{i,t}(\theta)\hat{A}_{i,t}, \text{clip}(\rho_{i,t}(\theta), 1-\epsilon, 1+\epsilon)\hat{A}_{i,t} \right) \right) \right] - \beta \mathbb{D}_{\text{KL}}[\pi_{\theta} || \pi_{\text{ref}}] $$}
where $\rho_{i,t}(\theta) = \frac{\pi_{\theta}(o_{i,t}|q,o_{i,<t})}{\pi_{\theta_{\text{old}}}(o_{i,t}|q,o_{i,<t})}$ is the probability ratio, $\pi_{\theta_{\text{old}}}$ is the policy from which outputs were sampled, $\pi_{\text{ref}}$ is a reference policy, $\epsilon$ is a clipping hyperparameter, and $\beta$ controls the KL divergence penalty. This formulation avoids training a separate value model, thereby reducing memory and computational overhead, and aligns well with reward models trained on comparative data.

\subsection{Limitations}
While our approach demonstrates strong generalization across diverse medical VQA tasks, its effectiveness is still constrained by the limited scope of publicly available medical datasets. Despite our efforts to construct the Med-Zero-17K dataset, existing data sources often lack coverage in rare conditions, underrepresented modalities, and detailed reasoning annotations. This scarcity may limit the diversity and complexity of reasoning patterns the model can learn. Future work could explore collaboration with clinical institutions to access broader, more representative datasets, and investigate synthetic data augmentation to further enhance training diversity.

\subsection{CoT Reasoning for Cold Start Training Construction}

\begin{figure}
\centering
\includegraphics[width=\linewidth]{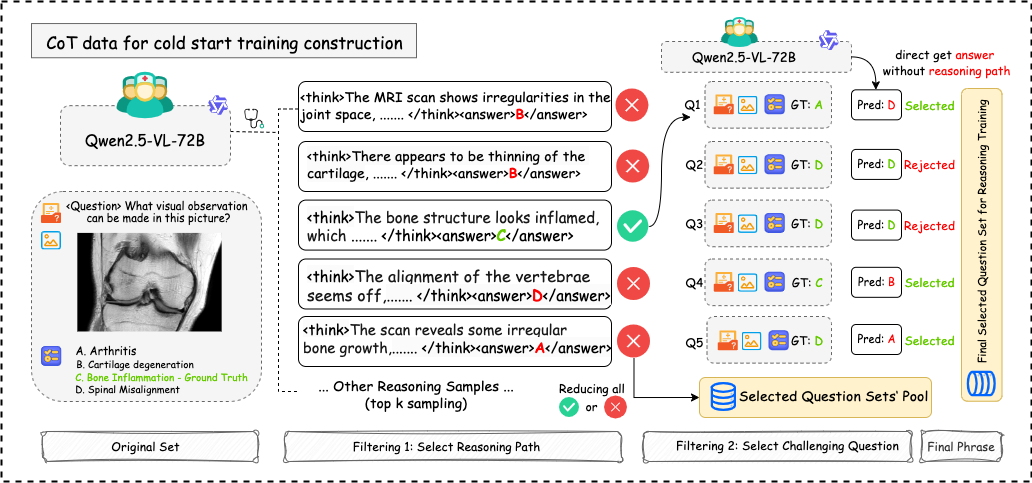}
\caption{CoT Cold Start Reasoning Data Construction}
\label{fig:cot}
\end{figure}

In this study, we explore the construction of cold start training using CoT data, aiming to enhance the reasoning capabilities of models in Med-VQA tasks.
Figure~\ref{fig:cot} illustrates the detailed steps of this process. Starting with an original set of medical images, a visual question is posed, such as "What visual observation can be made in this picture?" We then utilize the pre-trained Qwen2.5-VL-72B model to generate multiple reasoning paths. These paths include detailed analysis of the image and logical reasoning to reach a final answer. For instance, the model might identify irregularities in the joint space, thinning of the cartilage, signs of inflammation in the bone structure, misalignment of the vertebrae, or irregular bone growth.
After generating several reasoning paths, we select the most plausible ones through a filtering strategy. This involves two main filtering processes: first, we select reasoning paths that align with the correct answers; second, we further filter to select challenging questions that require more complex logic and deeper analysis during the reasoning process.
Ultimately, we extract a high-quality question set from the selected reasoning paths for subsequent reinforcement learning training. This process ensures that the model is exposed to high-quality reasoning examples from the outset and enhances the model's generalization capabilities through diverse reasoning paths.
\subsection{Broader Impacts}
Our work contributes to the advancement of medical AI by introducing a scalable reinforcement learning framework and a high-quality dataset tailored for medical visual question answering. By improving consistency and generalization in clinical reasoning, CAPO has the potential to assist healthcare professionals in diagnostic decision-making, particularly in resource-limited settings. However, we acknowledge that the current reliance on publicly available datasets may limit representation across demographic groups, rare diseases, and global health contexts. Additionally, while our system is not intended for direct clinical use, inappropriate deployment without expert oversight could pose risks. We encourage future work to incorporate fairness, transparency, and real-world validation in collaboration with clinical stakeholders to ensure safe and equitable deployment.
\subsection{Training Parameters}
We trained our model using 8 NVIDIA A100 Tensor Core GPU. Each rollout worker used a batch size of 128, with a global batch size of 128. The micro-batch sizes were set to 4 for policy updates and 16 for experience collection. The KL loss coefficient was 1e-2, and entropy regularization was set to 1e-3 to encourage exploration. We used a maximum prompt length of 25,600 tokens and a maximum response length of 4,096 tokens to accommodate long clinical contexts. Training was conducted with Ray and vLLM backend, leveraging the Qwen2.5-VL-7B model as the base.
\subsection{Prompt Template}
\begin{table}[h!]
\centering
\caption{Prompt Template for Consistency Review} 
\label{tab:prompt_template_consistency} 
\begin{tabularx}{\textwidth}{@{} l >{\raggedright\arraybackslash}X @{}}
\toprule
\textbf{Prompt Section} & \textbf{Content / Instruction} \\
\midrule

Initial Request & Please review the "Think" (thought process) and "Answer" provided below. Referring to the "Question" for context, determine if the "Think" and "Answer" are consistent. \\
\addlinespace 

Definition & "Consistent" means: The logical reasoning in the "Think" process can reasonably lead to the "Answer", and the "Answer" aligns with the final conclusion of the "Think" process. \\
\addlinespace

Input Placeholders & \texttt{Question: \{question\}} \\
                   & \texttt{Think: \{think\}} \\
                   & \texttt{Answer: \{answer\}} \\
\addlinespace

Output Logic & If they are consistent, please answer: \texttt{yes} \\
             & If they are inconsistent (e.g., the conclusion of the "Think" process contradicts the "Answer", or the "Answer" is not derived from the "Think" process), please answer: \texttt{no} \\
\addlinespace

Final Instruction & Now output your judgement with \texttt{yes} or \texttt{no} directly: \\
\bottomrule
\end{tabularx}
\end{table}

\subsection{3D VQA Performance Analysis}

\begin{figure}[htbp]
\centering
\includegraphics[width=0.5\linewidth]{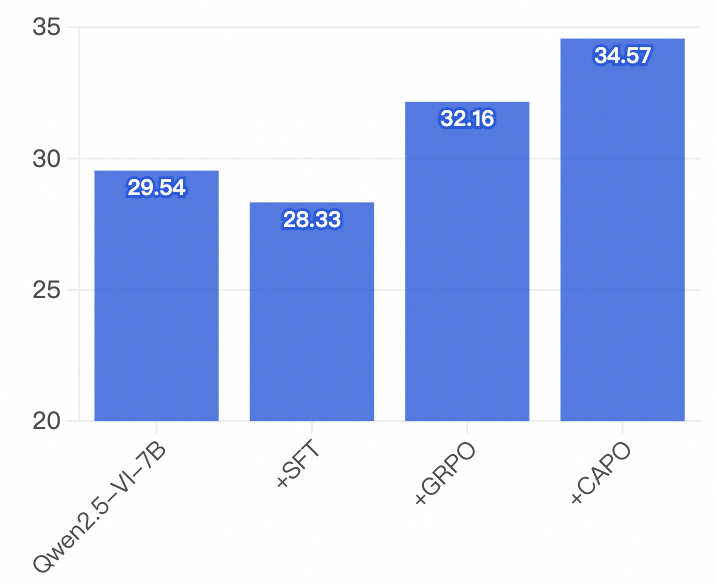} 
  \caption{3D VQA Performance.}
  \label{fig:3d_results}
\end{figure}

Figure~\ref{fig:3d_results} provides a comparative analysis of the performance of various training methodologies on 3D VQA tasks. The bar chart displays the accuracy percentages of the Qwen2.5-VL-72B model under different training conditions: without any additional training (base model), with Supervised Fine-Tuning (SFT), with Group Relative Policy Optimization (GRPO), and with our proposed Consistency-Aware Policy Optimization (CAPO).
The base Qwen2.5-VL-72B model, without further training, achieves a moderate accuracy of 29.54\%. When subjected to SFT, the model's performance slightly decreases to 28.33\%, indicating that this training approach may not be optimal for 3D VQA tasks. The application of GRPO leads to a more pronounced improvement, with the accuracy rising to 32.16\%, suggesting that this policy optimization method is more effective in enhancing the model's performance on these tasks.
Most significantly, the implementation of CAPO results in the highest accuracy of 34.57\%. This substantial increase over the other methods underscores the efficacy of CAPO in refining the model's reasoning and decision-making processes for 3D VQA. The enhanced performance is attributed to CAPO's unique approach of integrating rewards that promote consistency across different stages of the reasoning process, thereby fostering more reliable and coherent responses.

\subsection{Modality and Clinical Task Checklists of Med-Zero-17K}
\vspace{-1em}
\begin{table}[h!]
\centering
\small 
\label{fig:append_modality}
\caption{Clinical Task Composition of Med-Zero-17K}
\begin{tabularx}{\textwidth}{c|c|c}
\toprule
\multicolumn{3}{c}{\textbf{Clinical Task Checklist}} \\ \midrule
Disease Diagnosis & Severity Grading & Organ Recognition Abdomen \\
Surgical Instrument Recognition & Counting & Bone  \\
Organ Recognition Thorax & Organ Recognition Neck & Blood Vessels Recognition \\
Microorganism Recognition& Attribute Recognition & Cell Recognition \\
Surgeon Action Recognition & Organ Recognition Pelvic & Surgical Workflow Recognition \\
Image Quality Grading & Muscle & Nervous Tissue \\
Tumor Detection &  Lesion Localization & Medical Image Segmentation \\
Disease Progression Prediction & Anatomical Structure Measurement & Tissue Classification\\
\bottomrule
\end{tabularx}
\end{table}
\vspace{-1em}
\begin{table}[h!]
\centering
\small 
\label{fig:append_modality}
\caption{Modality Composition Composition of Med-Zero-17K}
\begin{tabularx}{\textwidth}{c|c|c}
\toprule
\multicolumn{3}{c}{\textbf{Modality Checklist}} \\ \midrule
Plain X-ray & Texture Characterization of Bone Radiograph & Mammography \\
CT & MRI & UltraSound  \\
Doppler Ultrasound & Echocardiography & Endoscopic Ultrasound \\
Endoscopy& Colposcopy & OCT \\
Optical Coherence Tomography & Fundus Photography & Fundoscopy \\
Adaptive Optics Ophthalmoscopy & Dermoscopy & Dermatoscopy \\
Histopathology & Histopathology Slide & Microscopy \\
Angiography& CT Angiography& PET Scan \\
Scintigraphy& Infrared Reflectance imaging & Medical Photography \\
Medical diagram & Graph/Chart & Electroencephalogram\\
\bottomrule
\end{tabularx}
\end{table}


\end{document}